\documentclass[10pt]{article} 
\usepackage[preprint]{rlc}

\usepackage{amssymb}            
\usepackage{mathtools}          
\usepackage{mathrsfs}           
\mathtoolsset{showonlyrefs}     
\usepackage{graphicx}           
\usepackage{subcaption}         
\usepackage[space]{grffile}     
\usepackage{url}                

\usepackage{enumerate, amsmath, amsthm, amssymb, algorithm, algorithmic, pifont, csquotes, dashrule, tikz, bbm, booktabs, bm, verbatim}

\definecolor{dblue}{RGB}{98, 140, 190}
\definecolor{dlblue}{RGB}{216, 235, 255}
\definecolor{dgreen}{RGB}{124, 155, 127}
\definecolor{dpink}{RGB}{207, 166, 208}
\definecolor{dyellow}{RGB}{255, 248, 199}
\definecolor{dgray}{RGB}{46, 49, 49}

\newcommand{\durl}[1]{\textcolor{dblue}{\underline{\url{#1}}}}

\newcommand{\mc}[1]{\mathcal{#1}}

\newcommand{\bE}{\mathbb{E}}

\newcommand{\bR}{\mathbb{R}}
\newcommand{\bP}{\mathbb{P}}
\newcommand{\bI}{\mathbb{I}}
\newcommand{\bH}{\mathbb{H}}
\newcommand{\bN}{\mathbb{N}}

\newcommand{\kl}[2]{D_{\mathrm{KL}}(#1\text{ }||\text{ }#2)}

\newcommand{\iid}{\overset{i.i.d.}{\sim}}

\newcommand{\ra}{\rightarrow}

\DeclareMathOperator*{\argmax}{arg\,max}

\newcounter{DaveDefCounter}
\newtheorem{fact}{Fact}

\let\classAND\AND
\let\AND\relax
\usepackage{algorithmic}

\let\AND\classAND
\AtBeginEnvironment{algorithmic}{\let\AND\algoAND}

\floatname{algorithm}{Algorithm}

\newif\ifsubmit
\submitfalse
\ifsubmit
\newcommand{\dnote}[1]{}
\newcommand{\bnote}[1]{}
\else
\newcommand{\dnote}[1]{\textcolor{blue}{Dilip: #1}}
\newcommand{\bnote}[1]{\textcolor{orange}{Ben: #1}}
\fi

\title{Satisficing Exploration for Deep Reinforcement Learning}

\author{Dilip Arumugam\\
  Department of Computer Science\\
  Stanford University\\
  \texttt{dilip@cs.stanford.edu}
  \And
  Saurabh Kumar\\
  Department of Computer Science\\
  Stanford University\\
  \texttt{szk@stanford.edu}
  \And
  Ramki Gummadi\\
  Google Research, Brain Team\\
  \texttt{gsrk@google.com}
  \And
  Benjamin Van Roy \\
  Department of Electrical Engineering \\
  Department of Management Science \& Engineering\\
  Stanford University\\
  \texttt{bvr@stanford.edu}
}

\begin{document}
\maketitle

\begin{abstract}
A default assumption in the design of reinforcement-learning algorithms is that a decision-making agent always explores to learn optimal behavior. In sufficiently complex environments that approach the vastness and scale of the real world, however, attaining optimal performance may in fact be an entirely intractable endeavor and an agent may seldom find itself in a position to complete the requisite exploration for identifying an optimal policy. Recent work has leveraged tools from information theory to design agents that deliberately forgo optimal solutions in favor of sufficiently-satisfying or \textit{satisficing} solutions, obtained through lossy compression. Notably, such agents may employ fundamentally different exploratory decisions to learn satisficing behaviors more efficiently than optimal ones that are more data intensive. While supported by a rigorous corroborating theory, the underlying algorithm relies on model-based planning, drastically limiting the compatibility of these ideas with function approximation and high-dimensional observations. In this work, we remedy this issue by extending an agent that directly represents uncertainty over the optimal value function allowing it to both bypass the need for model-based planning and to learn satisficing policies. We provide simple yet illustrative experiments that demonstrate how our algorithm enables deep reinforcement-learning agents to achieve satisficing behaviors. In keeping with previous work on this setting for multi-armed bandits, we additionally find that our algorithm is capable of synthesizing optimal behaviors, when feasible, more efficiently than its non-information-theoretic counterpart.
\end{abstract}


\section{Introduction}
\label{sec:intro}

In recent years, there has been a tectonic shift in the most celebrated successes of deep reinforcement learning that transcends the initial arcade games~\citep{tesauro1995temporal,bellemare2013arcade,mnih2015human,silver2016mastering,schrittwieser2020mastering} where the sub-field began in favor of real-world applications~\citep{dulac2021challenges}; an incomplete list of some notable examples includes classic robotic control problems~\citep{lillicrap2016continuous,schulman2016high,akkaya2019solving}, theorem provers~\citep{kaliszyk2018reinforcement}, molecular design~\citep{popova2018deep,zhou2019optimization}, superpressure balloon flight controllers~\citep{bellemare2020autonomous}, computer chip layout design~\citep{mirhoseini2021graph}, matrix multiplication algorithms~\citep{fawzi2022discovering}, and sophisticated dialogue agents~\citep{stiennon2020learning,ouyang2022training}.

While these advances are exciting, they also bring awareness to some of the harsh realities that real-world decision-making agents must inevitably face as they continue to proliferate and push the frontier into novel application areas. It is important to recognize that these agents are computationally bounded and, in many problems of interest, must contend with additional real-world constraints on time and other resources. As a concrete example of a scenario that behooves satisficing solutions over optimal ones, consider an environment designed for a sequential decision-making agent to autonomously complete tasks on the Internet~\citep{shi2017world,yao2022webshop}, such as making e-commerce purchases or querying for pieces of information. Given the wealth of knowledge
sources and vendors that can be accessed through the Internet, any one query or desired item for purchase will likely yield a minimum of hundreds, if not thousands, of potential results. As has been well known from years of research on provably-efficient exploration in reinforcement learning~\citep{kearns2002near,kakade2003sample,strehl2009reinforcement,jin2018q}, identifying the globally optimal solution technically requires sifting through this vast number of candidates from start to finish, lest the agent settle prematurely and miss out on the best purchase deal or most accurate answer for an input query. However, these environments~\citep{shi2017world,yao2022webshop} are incredibly rich, complex, and contain a wealth of information that likely exceeds the capacity of any one agent; pursuing optimal behaviors in such environments may no longer be a tractable endeavor as agents are forced to continually explore in order to obtain the potentially unbounded amount of information needed for synthesizing an optimal policy.  

Recent work by \citet{arumugam2022deciding} examines this capacity-limited setting and introduces a Bayesian reinforcement-learning algorithm for prioritizing exploration around an alternative, surrogate learning target~\citep{lu2023reinforcement}, which strikes a balance between being sufficiently simple to learn while also being suitably performant for the task at hand; crucially, they ground this procedure for learning \textit{satisficing} behaviors~\citep{simon1955behavioral,simon1956rational,newell1958elements,newell1972human,simon1982models} formally through lossy compression and rate-distortion theory~\citep{shannon1959coding}, resulting in a posterior-sampling algorithm that is capable of making fundamentally different exploratory choices than those of an agent purely seeking optimal behavior without regard for its own capacity constraints. One limitation of their proposed algorithm is that it is model-based, hindering its application to large-scale problems by the prerequisite of approximate, near-optimal planning. 

While progress on the open problem of high-dimensional model-based planning continues~\citep{wang2019benchmarking}, this paper develops a model-free approach inspired by a long line of work that performs Thompson sampling~\citep{thompson1933likelihood, russo2018tutorial} over the optimal action-value function~\citep{osband2016deep,osband2016deepthesis,osband2016generalization,osband2019deep,osband2023approximate}, bypassing the need for planning. We introduce an algorithm that amortizes the lossy compression of \citet{arumugam2021deciding} across the state space, allowing classic algorithms from the information theory community for computing the rate-distortion function to become viable~\citep{blahut1972computation,arimoto1972algorithm}. We demonstrate this precisely through computational experiments which showcase a single deep reinforcement-learning algorithm that can be parameterized to achieve a broad spectrum of satisficing solutions, which attain varying degrees of performance.

The paper is organized as follows: we outline the problem formulation in Section \ref{sec:prob_form} before presenting our deep reinforcement-learning algorithm in Section \ref{sec:approach}. In Section \ref{sec:experiments}, we outline the core hypothesis that underlies our empirical investigation before presenting the complementary experimental results. In the interest of space, all algorithms, an overview of related work, and additional details around empirical results are relegated to the appendix.

\section{Problem Formulation}
\label{sec:prob_form}

We formulate a sequential decision-making problem as an infinite-horizon, discounted Markov Decision Process (MDP)~\citep{bellman1957markovian,Puterman94} defined by $\mc{M} = \langle \mc{S}, \mc{A}, \rho, \mc{T}, \mu, \gamma \rangle$. Here $\mc{S}$ denotes a set of states, $\mc{A}$ is a set of actions, $\rho:\mc{S} \times \mc{A} \ra [0,1]$ is a deterministic reward function providing evaluative feedback signals (in the unit interval) to the agent, $\mc{T}:\mc{S} \times \mc{A} \ra \Delta(\mc{S})$ is a transition function prescribing distributions over next states, $\mu \in \Delta(\mc{S})$ is an initial state distribution, and $\gamma \in [0,1)$ is the discount factor communicating a preference for near-term versus long-term rewards. Beginning with an initial state $s_0 \sim \mu$, for each timestep $t \in \bN$, the agent observes the current state $s_t \in \mc{S}$, selects action $a_t \sim \pi(\cdot \mid s_t) \in \mc{A}$, enjoys a reward $r_t = \rho(s_t,a_t) \in [0,1]$, and transitions to the next state $s_{t+1} \sim \mc{T}(\cdot \mid s_t, a_t) \in \mc{S}$. 

A stationary, stochastic policy $\pi:\mc{S} \ra \Delta(\mc{A})$, encodes a pattern of behavior mapping individual states to distributions over possible actions whose overall performance in any MDP $\mc{M}$ when starting at state $s \in \mc{S}$ and taking action $a \in \mc{A}$ is assessed by its associated action-value function $Q^\pi(s,a) = \bE\left[\sum\limits_{t=0}^\infty \gamma^t \rho(s_{t},a_{t}) \bigm| s_0 = s, a_0 = a\right]$, where the expectation integrates over randomness in the action selections and transition dynamics. Taking the corresponding value function as $V^\pi(s) = \bE_{a \sim \pi(\cdot \mid s)}\left[Q^\pi(s,a)\right]$ and letting $\Pi \triangleq \{\mc{S} \ra \Delta(\mc{A})\}$ denote the set of all stochastic policies, we define the optimal policy $\pi^\star$ as achieving supremal value with $V^\star(s) = \sup\limits_{\pi \in \Pi} V^\pi(s) = \max\limits_{a^\star \in \mc{A}} Q^\star(s,a^\star)$ and $Q^\star(s,a) = \sup\limits_{\pi \in \Pi} Q^\pi(s,a)$ for all $s \in \mc{S}$ and $a \in \mc{A}$. As the agent interacts with the environment over the course of $K \in \bN$ episodes, we let $\tau_k = (s^{(k)}_1, a^{(k)}_1, r^{(k)}_1, \ldots)$ be the random variable denoting the trajectory experienced by the agent in the $k$th episode, for any $k \in [K]$. Meanwhile, $H_k = \{\tau_1,\tau_2,\ldots, \tau_{k-1}\} \in \mc{H}_k$ is the random variable representing the entire history of the agent's interaction within the environment at the start of the $k$th episode. Abstractly, a reinforcement-learning algorithm is a sequence of policies $\{\pi^{(k)}\}_{k \in [K]}$ where, for each episode $k \in [K]$, $\pi^{(k)}:\mc{H}_k \ra \Pi$ is a function of the current history $H_k$.

Throughout the paper, we will denote the entropy and conditional entropy conditioned upon a specific realization of an agent's history $H_k$, for some episode $k \in [K]$, as $\bH_k(X) \triangleq \bH(X \mid H_k = H_k)$ and $\bH_k(X \mid Y) \triangleq \bH_k(X \mid Y, H_k = H_k)$, for two arbitrary random variables $X$ and $Y$. This notation will also apply analogously to the mutual information $\bI_k(X;Y) \triangleq \bI(X;Y \mid H_k = H_k) = \bH_k(X) - \bH_k(X \mid Y) = \bH_k(Y) - \bH_k(Y \mid X),$ as well as the conditional mutual information $\bI_k(X;Y \mid Z) \triangleq \bI(X;Y \mid H_k = H_k, Z),$ given an arbitrary third random variable, $Z$. Note that their dependence on the realization of random history $H_k$ makes both $\bI_k(X;Y)$ and $\bI_k(X;Y \mid Z)$ random variables themselves. The traditional notion of conditional mutual information given the random variable $H_k$ arises by integrating over this randomness: 
$$\bE\left[\bI_k(X;Y)\right] = \bI(X;Y \mid H_k) \qquad \bE\left[\bI_k(X;Y \mid Z)\right] = \bI(X;Y \mid H_k,Z).$$
Additionally, we will also adopt a similar notation to express a conditional expectation given the random history $H_k$: $\bE_k\left[X\right] \triangleq \bE\left[X|H_k\right].$

\section{Satisficing with Randomized Value Functions}
\label{sec:approach}

In this section, we extend the preceding problem formulation to the Bayesian reinforcement learning setting used throughout this work. We then introduce a deep reinforcement-learning algorithm that yields satisficing solutions via lossy compression of the optimal action-value function, $Q^\star$.

\subsection{Randomized Value Functions}
\label{sec:rvf}

Our work operates in the Bayesian reinforcement learning~\citep{bellman1959adaptive,duff2002optimal,ghavamzadeh2015bayesian} setting, wherein the underlying MDP the agent interacts with is unknown and, therefore, a random variable. An agent's initial uncertainty in this unknown, true MDP $\mc{M}$ is reflected by a prior distribution $\bP(\mc{M} \in \cdot \mid H_1)$. A typical objective for this setting is to design a provably-efficient reinforcement-learning algorithm that incurs bounded Bayesian regret, which simply takes the traditional notion of regret and applies an outer expectation under the agent's prior to account for the unknown environment. Unlike prior work~\citep{russo2022satisficing,arumugam2021deciding,arumugam2022deciding}, which successfully leverages rate-distortion theory in this regard, the focus of the present work is to give rise to a practical agent design that is compatible with deep reinforcement learning; up to now, this has only been realized for multi-armed bandit problems~\citep{arumugam2021deciding,arumugam2021the}. 

In the absence of (or without regard for) any limitations on agent capacity, one fruitful strategy for addressing this setting involves reducing the agent's epistemic uncertainty~\citep{der2009aleatory} through Thompson sampling~\citep{thompson1933likelihood,russo2018tutorial}, resulting in the well-studied Posterior Sampling for Reinforcement Learning (PSRL) algorithm~\citep{strens2000bayesian,osband2013more,osband2014model,abbasi2014bayesian,agrawal2017optimistic,osband2017posterior,lu2019information} that enjoys rigorous theoretical guarantees for provably-efficient exploration. Despite this fact, PSRL poses a significant computational hurdle for deployment in large-scale environments as acting optimally with respect to a single posterior sample in each episode (per Thompson sampling) requires MDP planning for the optimal policy. Even while PSRL can still support efficient learning when the policy used in each episode is only near-optimal (see Algorithm 5.1 of \citep{osband2016deepthesis}), obtaining such a policy is itself a computationally-intensive procedure, especially in conjunction with function approximation~\citep{wang2019benchmarking}. 

Due to these challenges, recent work that combines principled, uncertainty-based exploration with deep reinforcement learning has been almost exclusively driven by the Randomized Value Functions (RVF) algorithm~\citep{osband2016generalization,osband2016deepthesis,o2018uncertainty,osband2019deep,osband2023approximate} which begins with a prior distribution over the optimal action-value function (rather than the MDP model $\langle \rho, \mc{T} \rangle$ as in PSRL) and, for each episode $k \in [K]$, performs Thompson sampling with respect to the current posterior distribution $\widehat{Q} \sim \bP(Q^\star \in \cdot \mid H_k)$. While maintaining an equivalence to PSRL in the tabular MDP setting (see Theorem 7.1 of \citep{osband2016deepthesis}) alongside a complementary regret bound, RVF avoids the computational inefficiencies of PSRL by reasoning over statistically-plausible value functions and, at each episode, behaving greedily with respect to the sampled action-value function: $\pi^{(k)}(s) \in \argmax\limits_{a \in \mc{A}} \widehat{Q}(s, a)$. 

While preliminary instantiations of RVF with deep neural networks relied on maintaining an approximate posterior over $Q^\star$ via computationally-inefficient ensembles~\citep{osband2016deep,osband2018randomized,dwaracherla2022ensembles} or hypernetworks~\citep{dwaracherla2020hypermodels}, recent follow-up work has gone on to address these limitations and retain the benefits of RVF with only a modest increase in computational effort~\citep{osband2021epistemic,osband2023approximate}.

\subsection{Blahut-Arimoto Randomized Value Functions}
\label{sec:blarvf}

Just as an agent employing PSRL will relentlessly explore to identify the underlying MDP $\mc{M}$~\citep{arumugam2022deciding}, a RVF agent will engage in a similar pursuit of the optimal action-value function $Q^\star$. While this reality reflects a desire to always act in search of optimal behavior, real-world reinforcement learning must instead contend with a simple, computationally-bounded agent interacting within an overwhelmingly-complex environment where limitations on time and resources may cause optimal behavior to no longer reside within the agent's means~\citep{arumugam2021deciding,russo2022satisficing,lu2023reinforcement,arumugam2022deciding}. A line of prior work~\citep{russo2017time,russo2018satisficing,arumugam2021deciding,arumugam2021the} has studied and addressed this issue for the multi-armed bandit setting~\citep{lai1985asymptotically,bubeck2012regret,lattimore2020bandit} while progress for reinforcement learning has been exclusively theoretical~\citep{arumugam2022deciding} in nature.

Rather than unrealistically presuming an agent has the (potentially unlimited) capacity needed to negotiate amongst these numerous choices and acquire the requisite bits of information for identifying the best option, one might instead take inspiration from human decision makers~\citep{tenenbaum2011grow,lake2017building}, whose cognition is known to be resource-limited~\citep{simon1956rational,newell1958elements,newell1972human,simon1982models,gigerenzer1996reasoning,vul2014one,griffiths2015rational,gershman2015computational,lieder2020resource,bhui2021resource,brown2022humans,ho2022people}, and settle for a near-optimal or satisficing solution. In this paper, we build upon a long-line of work in the cognitive-science literature that formalizes the limits of bounded decision-makers using the tools of information theory and doing so in tandem with reinforcement learning~\citep{sims2003implications,peng2005learning,parush2011dopaminergic,botvinick2015reinforcement,sims2016rate,sims2018efficient,zenon2019information,ho2020efficiency,gershman2020reward,gershman2020origin,mikhael2021rational,lai2021policy,gershman2021rational,jakob2022rate,bari2022undermatching,arumugam2024bayesian}. Crucially, however, we do so in a manner meant to retain and generalize the elegant theoretical properties of RVF while also maintaining its compatibility with deep reinforcement learning.

We design an agent that solves a rate-distortion optimization on a per-timestep basis once the current state has already been observed. Consequently, the learning target computed by each lossy compression problem is a target action $\tilde{A}_t$~\citep{arumugam2021deciding,arumugam2021the,arumugam2024bayesian} that leverages current knowledge of $Q^\star$ to achieve satisficing performance when executed from each state. Thus, for any state $s \in \mc{S}$ and distortion threshold $D \in \bR_{\geq 0}$, the resulting rate-distortion function is given by
$\mc{R}_k(s, D) = \inf\limits_{\widetilde{A} \in \mc{A}} \bI_k(Q^\star; \widetilde{A}) \text{ s.t. } \bE_k\left[d_{s}(Q^\star,\widetilde{A})\right] \leq D,$
where the distortion function $d_s: \mc{Q} \times \mc{A} \ra \bR_{\geq 0}$ induced by any state $s \in \mc{S}$ is defined as $d_s(Q^\star,\widetilde{a}) = \left(\max\limits_{a \in \mc{A}} Q^\star(s,a) - Q^\star(s,\widetilde{a})\right)^2.$

From an information-theoretic perspective, this formulation is akin to lossy source coding with side information available at the decoder~\citep{wyner1976rate,berger1998lossy}. Thinking about the extremes of this rate-distortion trade-off, notice that exclusive concern with rate minimization ($D \uparrow \infty$) yields a uniform distribution over all actions at each state; meanwhile, exclusive concern with distortion minimization ($D = 0$) recovers the greedy action for the particular realization of $Q^\star$, as RVF would. An agent only concerned with optimal behavior must obtain all $\bH_1(Q^\star)$ bits of information in order for each of the rate-distortion optimization problems to recover the optimal action at each state. In contrast, an agent that is only interested in target actions that are easy to learn obtains a uniform distribution over actions in every state. Naturally, the intermediate region between these extremes reflects a spectrum of satisficing policies that, within each state, focuses on learning a more tractable, near-optimal action in a manner analogous to satisficing algorithms in the multi-armed bandit setting~\citep{arumugam2021deciding,arumugam2021the}.

We may employ the classic Blahut-Arimoto algorithm~\citep{blahut1972computation,arimoto1972algorithm} to compute the channel achieving the rate-distortion limit at each timestep; rather than having an explicit distortion threshold $D$, this algorithm consumes as input a Lagrange multiplier $\beta \in \bR_{\geq 0}$ that communicates an implicit preference for the desired trade-off between rate and distortion. While only computationally feasible for a discrete information source and a discrete channel output, the latter requirement is immediately satisfied for MDPs with a discrete action space ($|\mc{A}| < \infty$) or a suitably-fine quantization of a continuous action space. To address the former constraint around the information source, we may employ the so-called ``plug-in estimator'' of the rate-distortion function~\citep{harrison2008estimation}, which replaces a continuous information source with the discrete empirical distribution obtained via Monte-Carlo sampling and is not only asymptotically consistent~\citep{harrison2008estimation} but also admits a finite-sample approximation guarantee~\citep{palaiyanur2008uniform}. 

The resulting Blahut-Arimoto Randomized Value Functions (BA-RVF) algorithm is given as Algorithm \ref{alg:ba_rvf} in Appendix \ref{sec:algs}. It is worth mentioning that BA-RVF does bear increased computational cost per-timestep in comparison to traditional RVF and future work may benefit from using ideas like distillation~\citep{rusu2015policy} to reduce these costs during rollouts in exchange for increased computational overhead between episodes. An additional drawback of BA-RVF is the dependence of agent performance on the hyperparameter, $\beta$; unfortunately, as $\beta$ is a Lagrange multiplier~\citep{boyd2004convex}, it must be tuned on a per-problem basis although future work may benefit from finding heuristic schemes for tuning or adapting $\beta$ over time that work well across a broad range of problems.

\begin{figure}
\centering
\begin{subfigure}{.45\textwidth}
  \centering
  \includegraphics[height=.35\linewidth]{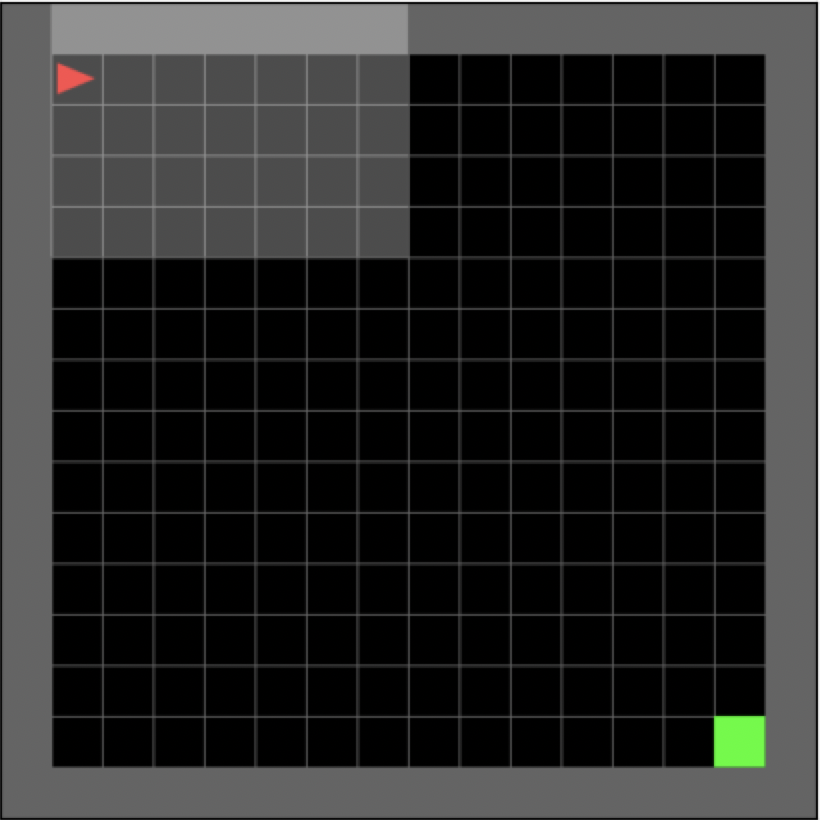}
  \includegraphics[height=.7\linewidth]{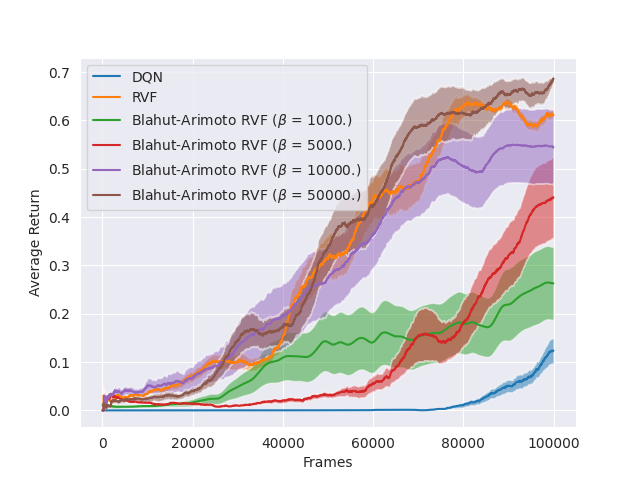}
  \caption{MiniGrid-Empty}
    \label{fig:empty_results}
\end{subfigure}
\begin{subfigure}{.45\textwidth}
  \centering
  \includegraphics[height=.35\linewidth]{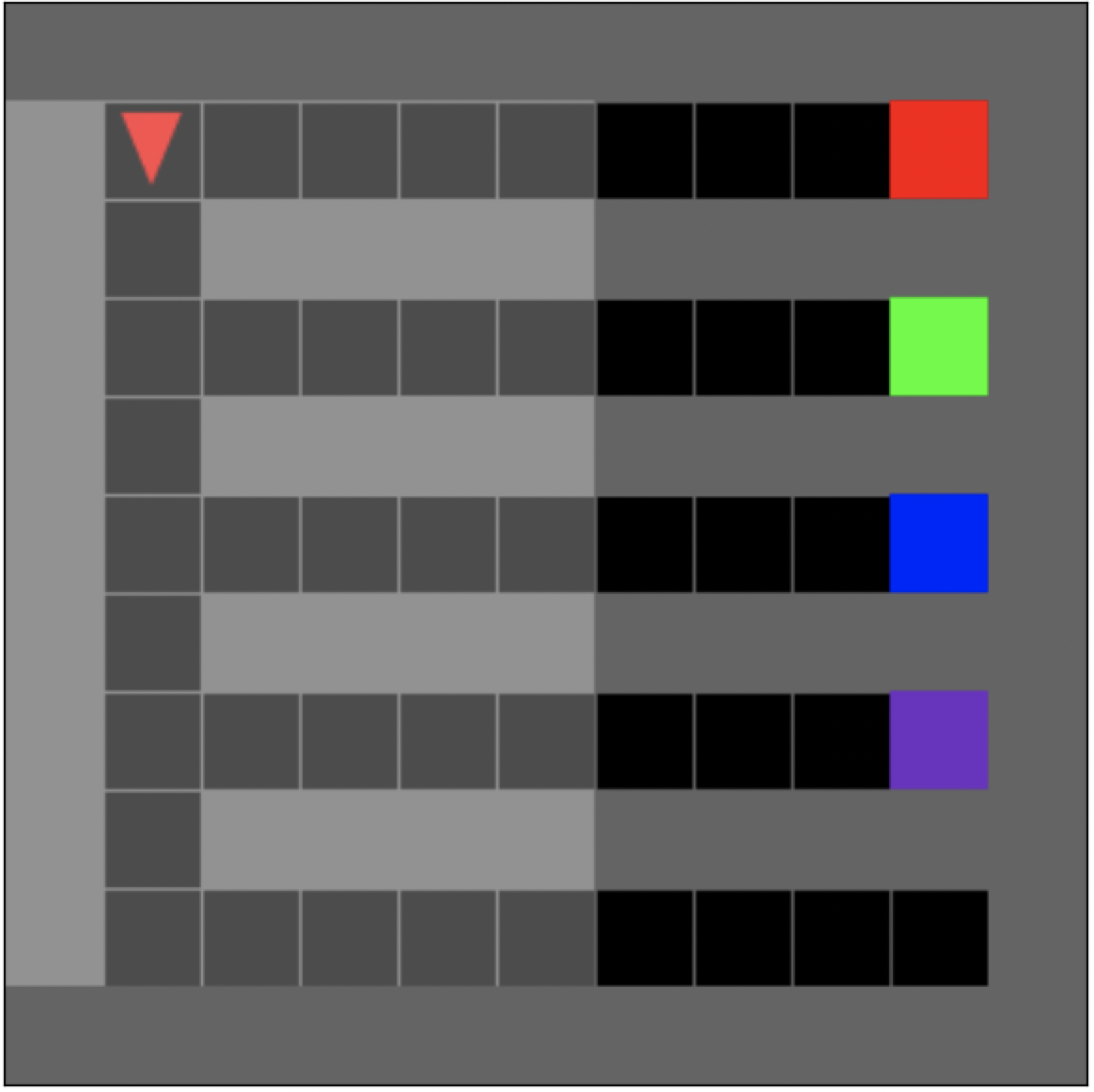}
  \centering
  \includegraphics[height=.7\linewidth]{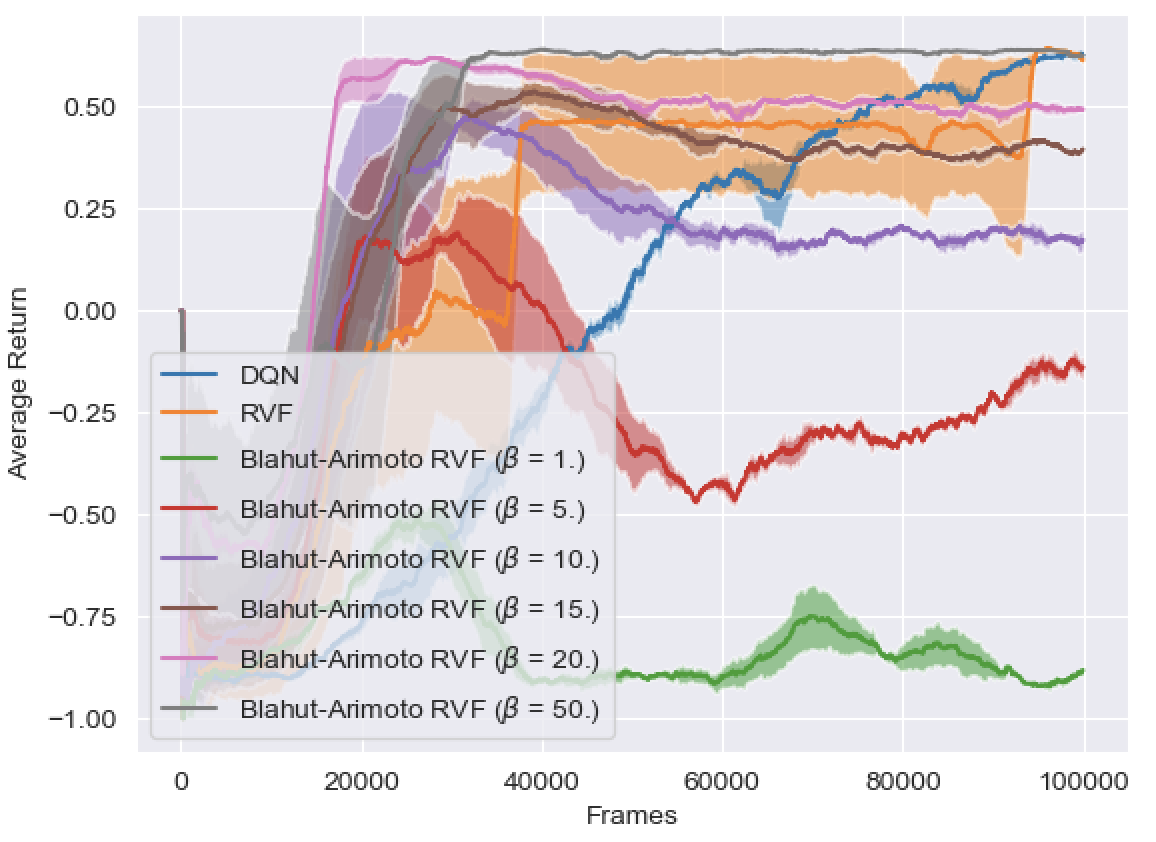}
  \caption{MiniGrid-CorridorEnv}
  \label{fig:corridor_results}
\end{subfigure}
\caption{(Top) MiniGrid environments used in our empirical evaluation of Blahut-Arimoto RVF. An observation is a partial image of the whole grid indicated by the shaded region. Black tiles represent empty squares, gray tiles represent walls, and colored tiles represent goal states. The agent begins in the upper left corner and an episode terminates when the agent either reaches a goal state or takes $100$ steps. (Bottom) Learning curves of DQN, RVF, and Blahut-Arimoto RVF.}
\label{fig:envs+results}
\end{figure}

\section{Experiments}\label{sec:experiments}

While the typical empirical evaluation for deep reinforcement-learning agents centers around demonstrating efficient acquisition of optimal behaviors, our experiments instead aim to elucidate how our proposed Blahut-Arimoto RVF algorithm (Algorithm \ref{alg:ba_rvf}) yields a successful generalization of the standard RVF algorithm capable of recovering a broad spectrum of satisficing solutions while still retaining the ability to gracefully address the challenge of exploration. To this end, we begin with results on two simple yet illustrative tasks that serve as unit tests of our core empirical hypothesis and leave the task of orchestrating a large-scale empirical demonstration of our algorithm to future work. In particular, we build our evaluation around two MiniGrid environments \cite{minigrid}: (1) MiniGrid-Empty-16x16-v0, a standard MiniGrid domain with a single terminal goal state providing sparse positive reward, and (2) MiniGrid-Corridor-v0, a custom designed environment containing multiple goal states which provide rewards proportional to their distance from the agent's initial position. As noted in Figure \ref{fig:envs+results}, both environments are partially observable where the agent is given a limited egocentric view of the world and has three movement actions available for execution: \textsc{RotateLeft}, \textsc{RotateRight}, and \textsc{Forward}.

In both domains, we train Blahut-Arimoto RVF agents with different values of the Lagrange multiplier $\beta \in \bR_{\geq 0}$ to verify that this parameter successfully controls the trade-off between rate and distortion; concretely, larger values of $\beta$ should yield agents more concerned with learning near-optimal policies. To contextualize the results achieved by each of these Blahut-Arimoto RVF agents, we also include baseline results attained by standard DQN~\citep{mnih2015human} and RVF agents, where the latter uses an epistemic neural network~\citep{osband2021epistemic} for representing uncertainty over $Q^\star$ in a computationally-efficient manner. We train all agents for $100,000$ frames and report the average un-discounted episodic return achieved throughout training over multiple independent trials (8 seeds on MiniGrid-Empty-16x16-v0 and 3 seeds on MiniGrid-CorridorEnv-v0).

\begin{figure}[H]
    \centering
    \includegraphics[width=.8\linewidth]{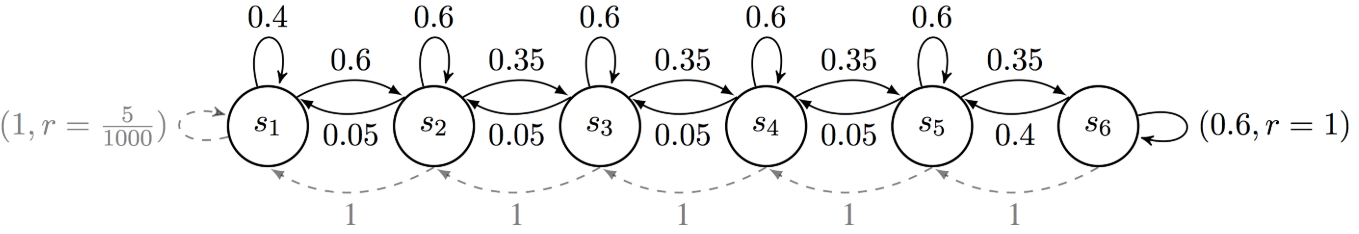}
    \caption{The RiverSwim MDP of \citet{strehl2008analysis} as studied by \citet{osband2013more}.}
    \label{fig:riverswim}
\end{figure}

In order to further underscore how our Blahut-Arimoto RVF agent achieves satisficing behaviors while retaining the strategic effectiveness of deep exploration, we offer an additional experiment using a variant of the classic RiverSwim environment~\citep{strehl2008analysis} show in Figure \ref{fig:riverswim} from \citet{osband2013more}.

Specifically, we provide results for an environment called ConfluenceSwim\footnote{A confluence is the point where multiple rivers meet.} which consists of three instances of the RiverSwim environment joined by a common initial state. Importantly, the ``current'' of each river that governs the success of the agent's movements towards the rewarding state at the opposite end varies and leads to three different levels of difficulty. The hardest level of difficulty adopts the transition structure shown in Figure \ref{fig:riverswim}
while the other rivers allow the agent to more easily swim upstream. Naturally, swimming up the rivers with weaker currents yields a smaller reward than engaging with the hardest river. Strategic deep exploration, however, is still needed to successfully traverse any one of the rivers.

\begin{figure}[H]
    \centering
    \includegraphics[width=.53\linewidth]{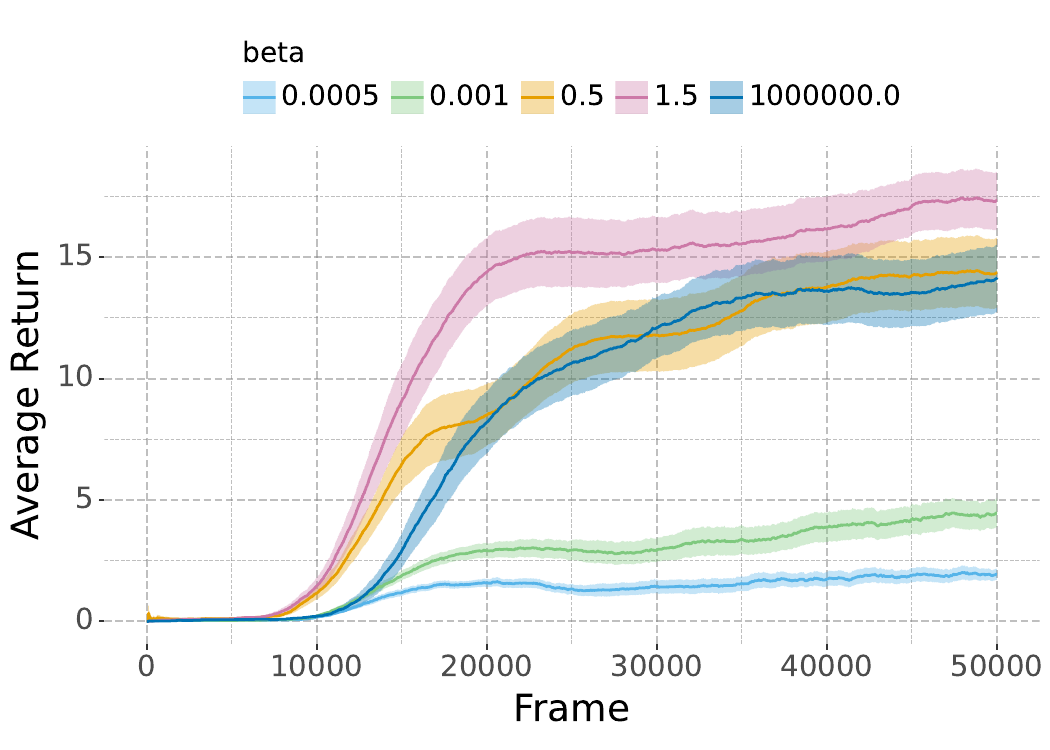}
    \caption{Learning curves for BA-RVF varying $\beta$ valuse in the ConfluenceSwim environment.}
    \label{fig:cswim_results}
\end{figure}

Figure \ref{fig:cswim_results} shows how varying the Lagrange multiplier hyperparameter in Blahut-Arimoto RVF still yields a spectrum of satisficing behaviors where the agent may opt to settle for the smaller reward at one of the easier rivers instead of traversing the most challenging river to obtain an optimal policy. Moreover, just as in prior work on rate-distortion-theoretic exploration for multi-armed bandit problems~\citep{arumugam2021deciding,arumugam2021the}, we observe the existence of a $\beta$ value for BA-RVF that synthesizes optimal behavior more efficiently than classic RVF, which can be interpreted as running BA-RVF with an incredibly large $\beta = 10^6$ such that the resulting target action at each state is the greedy action for a single, fixed posterior sample.

\section{Conclusion}

In this work, we challenge a core premise of agent design in deep reinforcement learning: that an agent should orient its exploration in pursuit of optimal behavior without regard for the complexity of the underlying environment. Using rate-distortion theory, we offer an agent designed to prioritize exploration towards satisficing behaviors and successfully dovetails with deep reinforcement learning. Our computational results demonstrate the efficacy of this agent in not only generalizing to accommodate satisficing solutions while retaining a graceful handling of the exploration challenge but also in synthesizing optimal solutions more efficiently than its non-satisficing counterpart. Future work still remains to precisely clarify how data efficiency factors into learning these satisficing behaviors.



\bibliography{references}
\bibliographystyle{rlc}

\appendix

\section{Preliminaries}
\label{sec:prelims}

In this section, we provide brief background on information theory, rate-distortion theory, and details on our notation. All random variables are defined on a probability space $(\Omega, \mc{F}, \bP)$. For any natural number $N \in \bN$, we denote the index set as $[N] \triangleq \{1,2,\ldots,N\}$. For any arbitrary set $\mc{X}$, $\Delta(\mc{X})$ denotes the set of all probability distributions with support on $\mc{X}$. For any two arbitrary sets $\mc{X}$ and $\mc{Y}$, we denote the class of all functions mapping from $\mc{X}$ to $\mc{Y}$ as $\{\mc{X} \ra \mc{Y}\} \triangleq \{f \mid f:\mc{X} \ra \mc{Y}\}$.


Here we introduce various concepts in probability theory and information theory used throughout this paper. We encourage readers to consult \citep{cover2012elements,gray2011entropy,duchi21ItLectNotes,polyanskiy2022IT} for more background. We define the mutual information between any two random variables $X,Y$ through the Kullback-Leibler (KL) divergence: $$\bI(X;Y) = \kl{\bP((X,Y) \in \cdot)}{\bP(X \in \cdot) \times \bP(Y \in \cdot)}, \qquad \kl{P}{Q} = \begin{cases} \int \log\left(\frac{dP}{dQ}\right) dP & P \ll Q \\ +\infty & P \not\ll Q \end{cases},$$ where $P$ and $Q$ are both probability measures on the same measurable space and $\frac{dP}{dQ}$ denotes the Radon-Nikodym derivative of $P$ with respect to $Q$. 
We define the entropy and conditional entropy for any two random variables $X,Y$ as $\bH(X) = \bI(X;X)$ and $\bH(Y \mid X) = \bH(Y) - \bI(X;Y)$, respectively. This yields the following identity for the conditional mutual information of any three arbitrary random variables $X$, $Y$, and $Z$: $\bI(X;Y|Z) = \bH(X|Z) - \bH(X \mid Y,Z) = \bH(Y|Z) - \bH(Y | X,Z).$
Through the chain rule of the KL-divergence, we obtain the chain rule of mutual information: $\bI(X;Y_1,\ldots,Y_n) = \sum\limits_{i=1}^n \bI(X;Y_i \mid Y_1,\ldots,Y_{i-1}).$


Here we offer a high-level overview of rate-distortion theory~\citep{shannon1959coding,berger1971rate} and encourage readers to consult \citep{cover2012elements} for more details. A lossy compression problem consumes as input a fixed information source $\bP(X \in \cdot)$ and a measurable distortion function $d: \mc{X} \times \mc{Z} \ra \bR_{\geq 0}$ which quantifies the loss of fidelity by using $Z$ in place of $X$. Then, for any $D \in \bR_{\geq 0}$, the rate-distortion function quantifies the fundamental limit of lossy compression as $$\mc{R}(D) = \inf\limits_{Z \in \mc{Z}} \bI(X;Z) \text{ such that } \bE\left[d(X,Z)\right] \leq D,$$ where the infimum is taken over all random variables $Z$ that incur bounded expected distortion, $\bE\left[d(X,Z)\right] \leq D$. Naturally, $\mc{R}(D)$ represents the minimum number of bits of information that must be retained from $X$ in order to achieve this bounded expected loss of fidelity and, conveniently, is well-defined for abstract information source and channel output random variables~\citep{csiszar1974extremum}. Moreover, the rate-distortion function has useful structural properties:
\begin{fact}
$\mc{R}(D)$ is a non-negative, convex, and non-increasing function of $D \in \bR_{\geq 0}$.
\label{fact:rdf_convex}
\end{fact}

\section{Algorithms}
\label{sec:algs}

Here we present the algorithm introduced in the main body of the paper.

\begin{algorithm}[H]
   \caption{Blahut-Arimoto Randomized Value Functions}
\begin{algorithmic}
   \STATE {\bfseries Input:} Prior $\bP(Q^\star \in \cdot \mid H_1)$, Lagrange multiplier $\beta \in \bR_{\geq 0}$, Posterior samples $Z \in \bN$
   \FOR{$k \in [K]$}
   \STATE Draw posterior sample $Q_1 \sim \bP(Q^\star \in \cdot \mid H_k)$
   \FOR{$t = 1,2,\ldots,$}
   \STATE Observe current state $s_t$ and draw samples $Q_2,\ldots,Q_{Z} \iid \bP(Q^\star \in \cdot \mid H_k)$
   \STATE Compute distortions $d_{s_t}(Q_z, \tilde{a})$, $\forall z \in [Z], \tilde{a} \in \mc{A}$
   \STATE Run Blahut-Arimoto algorithm with $\beta$ to compute $\bP(\tilde{A} \in \cdot \mid Q^\star)$ achieving $\mc{R}_k(s_t,D)$ limit
   \STATE Sample action $a_t$ from policy $\pi^{(k)}(a_t \mid s_t) = \bP(\tilde{A} = a_t \mid Q^\star = Q_1)$ \ENDFOR 
   \STATE $H_{k+1} = H_k \cup \{\tau_k\}$ and update posterior $\bP(Q^\star \in \cdot \mid H_{k+1})$
   \ENDFOR
\end{algorithmic}
 \label{alg:ba_rvf}
\end{algorithm}

\section{Related Work}
\label{sec:rw}

Overall, this paper touches upon two rich veins of prior work in the reinforcement-learning literature: (1) provably-efficient reinforcement learning and (2) information-theoretic  reinforcement learning. While there are numerous works that fall under each of these areas, we isolate a relevant, comprehensive subset below to allow for a suitably clear juxtaposition with our approach.

On the side of provably-efficient reinforcement learning, there are various approaches that have been developed over the course of the last two decades~\citep{kearns2002near,brafman2002r,kakade2003sample,auer2009near,bartlett2009regal,strehl2009reinforcement,jaksch2010near,osband2013more,dann2015sample,krishnamurthy2016pac,osband2017posterior,azar2017minimax,dann2017unifying,agrawal2017optimistic,jiang2017contextual,jin2018q,dann2018oracle,zanette2019tighter,du2019provably,sun2019model,agarwal2020flambe,agarwal2020pc,jin2020provably,dong2021simple,agarwal2021theory,jin2021bellman,foster2021statistical,lu2023reinforcement} which vary both along the type of analyses and guarantees (bounds on PAC-MDP sample complexity, regret, or iteration complexity) as well as the underlying structural assumptions leveraged to obtain those guarantees (tabular MDPs, linear function approximation, low-rank transition structure, bounded Bellman rank, \textit{etc.}). Due to our information-theoretic analysis, our results are quite general and can be applied to any MDP without structural assumptions such as finiteness of the state-action space. Additionally, our guarantees are obtained through a regret-analysis, although translations of these bounds to corresponding PAC-MDP bounds~\citep{kakade2003sample,strehl2009reinforcement} may be feasible~\citep{dann2017unifying,jin2018q}. 

Within this narrowed field of view, methods in this space can largely be segregated according to their use of optimism in the face of uncertainty or posterior sampling as the tool of choice for handling the exploration challenge, though recent work has considered blending ideas from both regimes~\citep{dann2021provably,agarwal2022model}; while both algorithm classes are theoretically sound and empirically effective, there are cases where the latter Bayesian reinforcement learning methods can be more favorable both in theory~\citep{osband2017posterior} as well as in practice~\citep{osband2016deep,osband2018randomized}. While some of the former optimism-based methods admit high-probability sample-complexity guarantees that depend on a sub-optimality parameter, the corresponding agents are designed to pursue optimal policies; in contrast, a core premise of this work is that an agent designer aware of a preference for satisficing behaviors over optimal ones can embed such considerations explicitly into the design of the agent which can, in turn, make fundamentally different exploratory choices during learning based on the objective of satisficing, rather than purely optimizing. 

Traditionally, the Bayesian reinforcement learning setting~\citep{bellman1959adaptive,duff2002optimal,ghavamzadeh2015bayesian} employed by posterior-sampling methods relies on the Bayes-Adaptive MDP (BAMDP) formulation~\citep{duff2002optimal}, which is notoriously intractable even in the tabular setting (see the discussion of \citet{arumugamplanning} for some conditions under which BAMDP planning may be resolved efficiently). The main innovation behind posterior-sampling methods~\citep{strens2000bayesian} is a lazy updating of the agent's epistemic uncertainty, removing the aforementioned intractability. Alternatively, one can consult \citet{ghavamzadeh2015bayesian} for details on methods that employ other approximation techniques or adhere to the PAC-BAMDP notion of efficient Bayesian reinforcement learning~\citep{kolter2009near,asmuth2009bayesian,sorg2010variance}. 

The core mechanism that underlies many posterior-sampling algorithms is Thompson sampling~\citep{thompson1933likelihood,russo2018tutorial} which, by design, is exclusively focused on obtaining optimal solutions. Even subsequent improvements to Thompson sampling that account for information gain~\citep{russo2014learning,russo2018learning} also retain this property. \citet{lu2023reinforcement} introduce the idea of a \textit{learning target} as a mechanism through which an agent may prioritize its exploration when the underlying environment is too immense and complex for the agent to be endlessly curious and pursue all available bits of information; crucially, however, their regret bounds presume that such a learning target has been computed \textit{a priori}. In contrast, the agents discussed in this work adaptively compute and refine the learning target as the agent's knowledge of the underlying environment accumulates. Most related to the present work are the algorithms of \citet{arumugam2021deciding,arumugam2021the} and \citet{arumugam2022deciding}, where the former algorithms employ learning targets adaptively computed via rate-distortion theory exclusively to multi-armed bandit problems while the latter translates these ideas over to PSRL, but without a concrete roadmap to a computationally-feasible instantiation. Our work remedies this final issue by maintaining and resolving epistemic uncertainty over the optimal action-value function, rather than the underlying model of the unknown MDP.

Setting aside work on provably-efficient reinforcement learning with guarantees obtained via information-theoretic analyses, the topic of information-theoretic reinforcement learning is largely an empirical body of work, focusing on how information-theoretic quantities may be practically applied by decision-making agents to address the fundamental challenges of generalization, exploration, and credit assignment. As the algorithms of this paper are primarily designed to address the challenge of exploration, we only mention in passing that work coupling information theory to address the challenges of temporal credit assignment is nascent~\citep{van2011hierarchical,arumugam2021information}. Meanwhile, there is a substantial literature on information-theoretic methods to aid generalization, largely situated around the information bottleneck principle~\citep{tishby2000information}, which instantiates a particular rate-distortion optimization to formalize the notion of a learned data representation that is both maximally compressive while also retaining the requisite information needed for task performance. This perspective leads to a information-theoretic formulation of classic state abstraction in reinforcement learning~\citep{li2006towards,van2006performance,abel2016near,abel2020thesis} for lossy compression of the original MDP state space~\citep{liu2011compressive,shafieepoorfard2016rationally,lerch2018policy,lerch2019rate,abel2019state}. A similar but distinct problem also manifests in the partially-observable MDP (POMDP)~\citep{aastrom1965optimal,kaelbling1998planning} setting where classic work in control theory models observational capacity limitations as an information-theoretic rate constraint~\citep{witsenhausen1971separation,mitter1999information,mitter2001control,borkar2001markov,crutchfield2001synchronizing,poupart2002value,tatikonda2004control,kostina2019rate} and asks how well one can control a system subject to such a rate limit.

Aligned with the perspective of this work, the bulk of the information-theoretic reinforcement-learning literature is aimed at addressing the challenge of exploration, logically expecting that novel information can serve as a useful form of intrinsic motivation~\citep{chentanez2004intrinsically,singh2009rewards} to guide the agent~\citep{storck1995reinforcement,polani2001information,iwata2004new,klyubin2005empowerment,todorov2007linearly,klyubin2008keep,still2009information,polani2009information,ortega2011information,van2011look,still2012information,tishby2011information,sun2011planning,rubin2012trading,ortega2013thermodynamics,mohamed2015variational,houthooft2016vime,goyal2018infobot,goyal2020variational}. Most notable among these methods are those that have inspired popular modern deep reinforcement-learning algorithms through the ``control as inference'' or KL-regularized reinforcement learning perspective~\citep{,toussaint2009robot,kappen2012optimal,levine2018reinforcement,ziebart2010modeling,fox2016taming,haarnoja2017reinforcement,haarnoja2018soft,galashov2019information,tirumala2019exploiting}. We refer readers to the short survey of \citet{arumugam2022rate} for an overview of how the principled Bayesian methods used in this work compare relative to these latter methods inspired by the  information bottleneck principle though, in short, the fundamental issue boils down to a lack of properly alleviating the burdens of exploration~\citep{o2020making}; that said, guarantees do exist for such maximum entropy exploration schemes in the absence of reward~\citep{hazan2019provably} although, translated into our real-world reinforcement learning setting where agents are fundamentally bounded, naively attempting to maximize entropy would be an ill-defined objective as all bits of information that could be acquired from the environment cannot be retained within the agent's capacity limitations, by assumption.

Finally, we conclude by noting that the practical deep reinforcement-learning algorithm developed in this work relies on the classic Blahut-Arimoto algorithm~\citep{blahut1972computation,arimoto1972algorithm} for computing the rate-distortion function when both the information source and channel output random variables are discrete. While the algorithm is known to be theoretically-sound in general~\citep{csiszar1974extremum} and globally-convergent~\citep{csiszar1974computation} under these conditions, various techniques have been developed in order to accelerate the Blahut-Arimoto algorithm and make it applicable to continuous information sources~\citep{boukris1973upper,rose1994mapping,sayir2000iterating,matz2004information,chiang2004geometric,dauwels2005numerical,niesen2007adaptive,vontobel2008generalization,naja2009geometrical,yu2010squeezing}. While we do not explore any of these extensions here, the reality that our proposed deep reinforcement-learning agent runs the Blahut-Arimoto algorithm on a per-timestep basis suggests that these works could serve as a useful basis for future work which more carefully studies large-scale deployment of our agent.

\section{Additional Minigrid Experiment Details}\label{sec:expdetails}

\subsection*{Algorithm Implementations}

\begin{table}[H]
\centering
\begin{tabular}{ |p{5cm}||p{5cm}|  }
 \hline
 \multicolumn{2}{|c|}{Hyperparameters for DQN} \\
 \hline
 Parameter & Value\\
 \hline
 discount ($\gamma$) & $0.99$ \\
 number of frames & $100,000$ \\
 optimizer   & Adam\\
 learning rate & $5 \cdot 10^{-4}$ \\
 replay buffer capacity & $25,000$ \\
 $\tau$ (used for soft target net updates) & $1 \cdot 10^{-3}$ \\
 batch size & $128$ \\
 max $\epsilon$ & $1.0$ \\
 min $\epsilon$ & $0.0$ \\
 warmup & $100$ frames \\ 
 $\epsilon$ decay rate & $\frac{1}{0.95 \cdot (\text{number of frames} - \text{warmup})}$ \\
 \hline
\end{tabular}
\caption{DQN Hyperparameters.}
\label{tab:dqn}
\end{table}

\begin{table}[H]
\centering
\begin{tabular}{ |p{5cm}||p{5cm}|  }
 \hline
 \multicolumn{2}{|c|}{Hyperparameters for RVF} \\
 \hline
 Parameter & Value\\
 \hline
 discount ($\gamma$) & $0.99$ \\
 number of frames & $100,000$ \\
 optimizer   & Adam\\
 learning rate & $5 \cdot 10^{-4}$ \\
 replay buffer capacity & $25,000$ \\
 $\tau$ (used for soft target net updates) & $1 \cdot 10^{-3}$ \\
 batch size & $128$ \\
 index dimension & $30$ \\
 noise scale & $0.1$ \\
 prior scale & $0.1$ \\
 \hline
\end{tabular}
\caption{RVF Hyperparameters.}
\label{tab:rvf}
\end{table}

\subsubsection*{DQN}
We implement the DQN agent as a convolutional neural network with the following architecture:
\begin{enumerate}
    \item Conv Layer: Kernel Size = $2 \times 2$, output channels = $16$, RELU activation
    \item Max Pool $2 \times 2$
    \item Conv Layer: Kernel Size = $2 \times 2$, output channels = $16$, RELU activation
    \item Conv Layer: Kernel Size = $2 \times 2$, output channels = $64$, RELU activation
    \item Fully-connected layer, output dimension = $128$, RELU activation
    \item Fully-connected layer, output dimension = $3$ (number of actions)
\end{enumerate}

\subsubsection*{RVF \& Blahut-Arimoto RVF}
We implement RVF, and consequently Blahut-Arimoto RVF, as an epistemic neural network comprised of a base network and an epinet \cite{osband2021epistemic}. For the base network, we use the same architecture as the DQN network. The epinet consists of a learnable network and a prior network and we guide our design choices around both components based on details reported by \citet{osband2021epistemic}. 

In order to reliably induce a distribution over functions, an epinet operates with an epistemic index $z \sim \mc{N}(0, I_d) \in \bR^d$, where the index dimension $d \in \bN$ is a hyperparameter. Intuitively, it is the stochasticity in the underlying reference distribution of $z$ that produces the stochasticity needed to represent a full distribution over functions. Both the learnable network and the prior network take as input a single sampled epistemic index concatenated with the embedding obtained by the convolutional layers of the base network as well as the penultimate layer activations of the base network. These last two components are first passed through a stop-gradient layer to avoid propagation of epinet gradients into the base network. 

For a finite action space $|\mc{A}| < \infty$, both the base network and the learnable network produce outputs in $\bR^{|\mc{A}|}$. Meanwhile, the prior network is itself a small ensemble of MLPs where the size of the ensemble is directly determined by the index dimension $d$. Consequently, the output of the entire prior network is determined by first feeding the inputs through all elements of the ensemble, stacking them together into a $|\mc{A}| \times d$ matrix, and then finally taking the matrix-vector product with the epistemic index $z \in \bR^d$ to obtain the action-values in $\bR^{|\mc{A}|}$. The base network and learnable network output vectors are immediately added together along with the prior network output after it has been multiplied by a prior scale parameter. 

As noted in \citet{osband2018randomized,dwaracherla2022ensembles}, the prior scale hyperparameter effectively controls the effective width or support of the initial distribution over value functions; large settings of this parameter will yield more variance between functions sampled through different epistemic indices but, simultaneously, will require prolonged interaction and stochastic gradient updates for the learnable network to compensate for the scaled prior effect. Our loss function for optimizing the epinet parameters consists of the standard TD(0)-error accompanied by Gaussian noise~\citep{osband2016deep,osband2019deep} where the standard deviation of the zero-mean noise is also a hyperparameter, denoted as the noise scale, which aligns with Bayesian linear regression. The resulting additive noise bonus is then computed during loss computations by taking the inner product between sampled noise variables and the epistemic indices in each minibatch. We encourage readers to consult \citep{osband2019deep,osband2021epistemic} for more details on epinets and inducing a distribution over optimal action-value functions in this manner.

 We use the following prior network architecture:
\begin{enumerate}
\item Fully-connected layer: output dimension = $5$, RELU activation
\item Fully-connected layer: output dimension = $5$, RELU activation
\item Fully-connected layer: output dimension = $3$ (number of actions)
\end{enumerate}

The learnable network architecture is as follows:
\begin{enumerate}
\item Fully-connected layer: output dimension = $256$, RELU activation
\item Fully-connected layer: output dimension = $256$, RELU activation
\item Fully-connected layer: output dimension = $3 \cdot 5$ (number of actions $*$ index dimension)
\end{enumerate}

The output from the epistemic network is computed as
$$
\text{base network output} + \text{learnable epinet output} + \text{prior scale} \cdot \text{prior network output}
$$

\subsection*{Hyper-parameter Selection}

On each environment, we performed a grid search over the prior scale and noise scale hyper-parameters of the RVF agent using a single seed. Intuitively, tuning these parameters is commensurate with ensuring that the true optimal action-value function of each domain lies in the support of the prior distribution represented by the corresponding epinet and that there is enough signal to facilitate deep exploration throughout the domain without compromising convergence. Specifically we performed a grid search over all combinations of noise scale $\in \{ 0.1, 0.15, 0.2 \}$ and prior scale $\in \{ 0.025, 0.05, 0.1, 0.25 \}$. On both domains, we found noise scale = $0.1$ and prior scale = $0.1$ to yield the highest average return after $100,000$ frames. As these hyperparameters govern the quality of how epistemic uncertainty is represented, but not how each algorithm utilizes that uncertainty to resolve exploration, we keep these tuned values fixed for all RVF and Blahut-Arimoto RVF experiments. 

Blahut-Arimoto RVF consumes as input the Lagrange multiplier $\beta \in \bR_{\geq 0}$ which controls the trade-off between rate and distortion. To determine which range of $\beta$ values covers the spectrum of minimizing rate to minimizing distortion, which can differ between environments, we first trained Blahut-Arimoto RVF with $\beta \in \{1, 10, 100, 1000, 10000, 100000, 1000000\}$. On MiniGrid-Empty-16x16-v0, the performance of Blahut-Arimoto RVF with $\beta = 1000000$ surpassed that of RVF, whereas on  MiniGrid-CorridorEnv-v0 this occured with a much smaller value of $\beta = 100$. Consequently, we used $1000000$ and $100$ as the upper bound on $\beta$ values to select for training Blahut-Arimoto RVF on MiniGrid-Empty-16x16-v0 and MiniGrid-CorridorEnv-v0, respectively.

Further hyper-parameter details for all agents are in Tables \ref{tab:dqn} and \ref{tab:rvf}.

\section{Compute Details}

For our experiments, we used an n1-standard-8 Google Cloud Virtual Machine with 1 NVIDIA Tesla P100 GPU.

\end{document}